\documentclass[letterpaper, 10 pt, conference]{ieeeconf}  
\usepackage{array}
\usepackage{times}  
\usepackage{helvet}  
\usepackage{courier}  
\usepackage[hyphens]{url}  
\usepackage{graphicx} 
\usepackage{multirow}
\usepackage{booktabs}
\usepackage[table]{xcolor} 
\usepackage{xcolor}
\usepackage{caption} 
\usepackage{amsthm}
\usepackage{algorithm}
\usepackage{algorithmic}
\usepackage{adjustbox}
\usepackage{newfloat}
\usepackage{listings}
\usepackage{amsfonts}
\usepackage{makecell}
\usepackage{float}
\usepackage{amsmath}
\usepackage{amssymb}
\usepackage{mathtools}
\usepackage{bm}
\usepackage[capitalize,noabbrev]{cleveref}

\theoremstyle{plain}

\theoremstyle{definition}

\theoremstyle{remark}

\newcommand{\method}{PD-VLA}

\IEEEoverridecommandlockouts                              

\title{\LARGE \bf
PD-VLA: Accelerating Vision-Language-Action Model Integrated with Action Chunking via Parallel Decoding
}

\author{ 
Wenxuan Song\textsuperscript{*1},
Jiayi Chen\textsuperscript{*1,5},
Pengxiang Ding\textsuperscript{2,3},
Han Zhao\textsuperscript{2,3},
Wei Zhao\textsuperscript{2},
Zhide Zhong\textsuperscript{1},\\
Zongyuan Ge\textsuperscript{4},
Zhijun Li\textsuperscript{5},
Donglin Wang\textsuperscript{2},
Lujia Wang\textsuperscript{1},
Jun Ma\textsuperscript{1},
Haoang Li\textsuperscript{1}}
\begin{document}

\twocolumn[{%
\renewcommand\twocolumn[1][]{#1}%
\maketitle
\begin{center}
    \centering
    \captionsetup{type=figure}
    \includegraphics[width=0.99\linewidth]{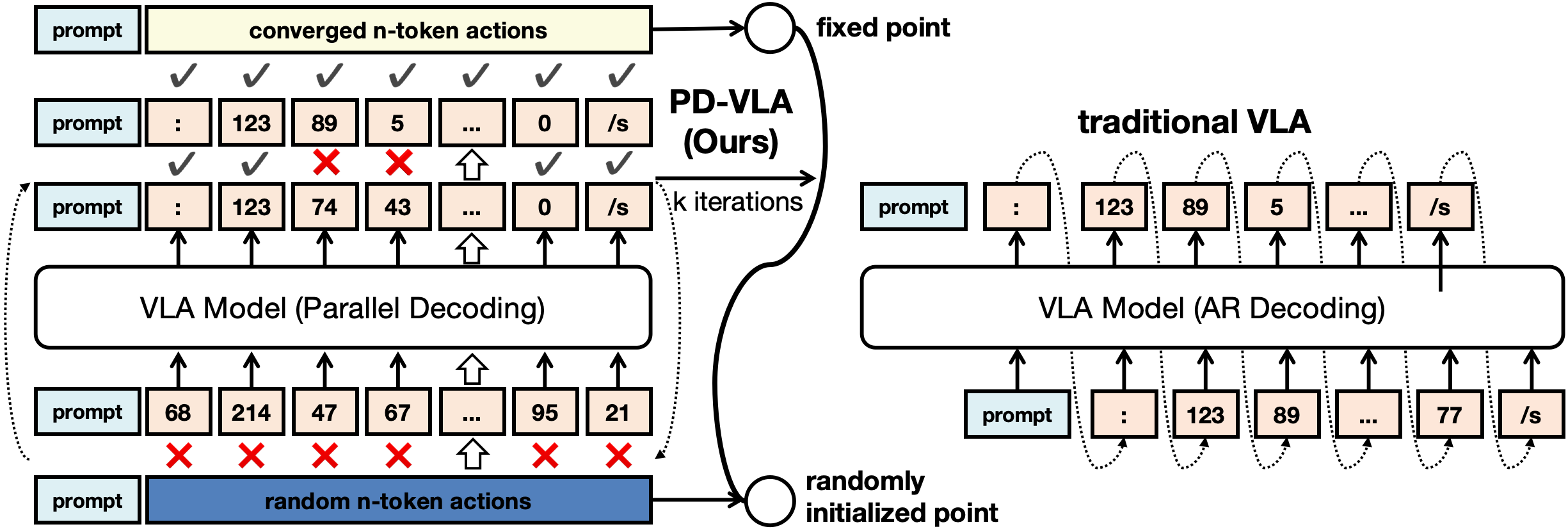}
    \captionsetup{font=footnotesize}
    \captionof{figure}{Comparison between the proposed parallel decoding on the left and traditional autoregressive (AR) decoding on the right. 
    Unlike AR decoding, which predicts action tokens sequentially, our parallel decoding simultaneously predicts all the tokens in parallel.}
    \label{fig:teaser}
    \vspace{1.5em}
\end{center}
}]

\begin{abstract}
Vision-Language-Action (VLA) models demonstrate remarkable potential for generalizable robotic manipulation.
The performance of VLA models can be 
improved by integrating with action chunking, a critical technique for effective control.
However, action chunking linearly scales up action dimensions in VLA models with increased chunking sizes. 
This reduces the inference efficiency.
Therefore, accelerating VLA integrated with action chunking is an urgent need.
To tackle this problem, we propose \textbf{\method}, the first parallel decoding framework for VLA models integrated with action chunking.
Our framework reformulates autoregressive decoding as a nonlinear system solved by parallel fixed-point iterations.
This approach preserves model performance with mathematical guarantees while significantly improving decoding speed. 
In addition, it enables training-free acceleration without architectural changes, as well as seamless synergy with existing acceleration techniques.
Extensive simulations validate that our \method~maintains competitive success rates while achieving 2.52$\times$ execution frequency on manipulators  (with 7 degrees of freedom) compared with the fundamental VLA model. 
Furthermore, we experimentally identify the most effective settings for acceleration.
Finally, real-world experiments validate its high applicability across different tasks.

\end{abstract}
\renewcommand{\thefootnote}{}
\noindent
\footnotetext{
This work was supported in part by the National Natural Science Foundation of China under Grant 62403401, in part by the Guangdong Basic and Applied Basic Research Foundation under Grant 2024A1515011992, in part by the Department of Education of Guangdong Province under Grant 2024KQNCX030, in part by the Guangzhou Municipal Education Project under Grant 2024312104, and in part by the Guangzhou-HKUST (GZ) Joint Funding Program under Grant 2025A03J3716. Corresponding Author: Haoang Li (haoangli@hkust-gz.edu.cn)

*Wenxuan Song and Jiayi Chen contributed equally to this work. 

\textsuperscript{1}The Hong Kong University of Science and Technology (Guangzhou), Guangzhou, China.

\textsuperscript{2}Westlake University, Hangzhou, China.

\textsuperscript{3}Zhejiang University, Hangzhou, China.

\textsuperscript{4}Monash University, Melbourne, Australia.

\textsuperscript{5}Harbin Institute of Technology, Harbin, China.}
\section{INTRODUCTION}

The pursuit of robust and generalizable robotic manipulation policies remains a fundamental challenge in embodied AI research~\cite{kim2020domain}. 
Recent advancements in Vision-Language Models (VLMs)~\cite{awadalla2023openflamingo,liu2024visual} have showcased impressive multimodal understanding capabilities, inspiring the development of Vision-Language-Action (VLA) models~\cite{rt1,rt2, octo_2023,niu2024llarva,song2024germ,kim24openvla}. These end-to-end architectures, which are trained on large-scale robotic datasets~\cite{o2024open,fang2024rh20t}, integrate visual perception and language understanding to directly generate executable actions.
This emerging paradigm shows strong effectiveness and generalization in diverse scenarios.

Recent VLA work~\cite{black2024pi_0, fast, vlas} has explored the integration with action chunking~\cite{act}, which highly improves the performance of VLA models in laboratory scenarios.
However, action chunking dramatically increases the action dimensions in a single inference.
For typical manipulators with 7 degrees of freedom (DoF)  (including 3-DoF translation, 3-DoF rotation, 1-DoF gripper), an action chunk of $m$ steps creates $7m$-dimensional action sequences.
This linearly increases single-inference time 
when autoregressive (AR) decoding is employed in VLA models.
The reason is that AR decoding sequentially predicts each token in an one-by-one manner.
As a result, the generation time is proportional to the predicted token length.
Therefore, there is an urgent need to accelerate the decoding process for VLA models integrated with action chunking.



\begin{table}[t]
\captionsetup{font=footnotesize}
\caption{Comparison between different acceleration methods for VLA models. ``Model-redesign-free'' indicates that a method does not redesign the foundation models. ``Training-free'' indicates that a method does not need training. ``Modification-free'' indicates that a method does not require modifications or adding auxiliary components to pre-trained VLA models. }
\small
\setlength\tabcolsep{2pt}
\label{tab:1}
\begin{tabular}{cccc}
\toprule  
\multirow{2}{*}{Methods} &  Model-redesign  & Training- & Modification- \\ 
 &  free   & free  & free\\  

\midrule  
TinyVLA \cite{wen2024tinyvla} & $\times$  & -  & - \\
RoboMamba \cite{liu2024robomamba} & $\times$   & -  & -  \\
\midrule  
QAIL \cite{park2024quantization} & \checkmark    & $\times$  & $\times$ \\
DeeR-VLA \cite{DeeR-VLA} & \checkmark  & $\times$   & $\times$ \\
\midrule  
VLA w/ Sparse.\cite{zhang2024sparsevlm} & \checkmark  & \checkmark   & $\times$\\
VLA w/ FastV \cite{fastv} & \checkmark  & \checkmark   & $\times$\\
VLA-Cache \cite{vlacache} & \checkmark  & \checkmark   & $\times$\\
\midrule 
\method~(ours) & \checkmark & \checkmark    & \checkmark \\
\bottomrule 
\end{tabular}
\end{table}

To address the above challenges, we present a novel parallel decoding framework for the mainstream VLA model with action chunking, called \textbf{P}arallel \textbf{D}ecoding for \textbf{VLA} (\textbf{\method}). 
Fig.~\ref{fig:teaser} illustrates the core concept of our parallel decoding approach.
Our key insight reframes AR action decoding as a system of nonlinear equations solved through parallel fixed-point iteration methods, e.g., Jacobi fix-point iteration
method~\cite{ortega2000iterative}. 
This approach preserves model performance with mathematical guarantees while significantly improving decoding speed. 
Please note that we only accelerate the decoding process during VLA inference.
Accordingly, our method enables 
friendly deployment, compared with existing methods, i.e., it achieves training-free acceleration without redesign and modification of models (see Table~\ref{tab:1}).
Moreover, our method achieves
seamless synergy with existing acceleration techniques.

We validate our \method~in extensive simulation and real-world experiments.
In simulation experiments, our method achieves significant acceleration without compromising performance.
Compared to the fundamental VLA model, our \method~achieves 2.52$\times$ execution frequency.
Furthermore, we experimentally identify the most effective settings for acceleration.
Finally, the real-world experiments show the strong applicability of \method, especially in the dexterous tasks, such as pouring the water.

Our primary contributions include:
\begin{itemize}
\item We propose the first parallel decoding framework for VLA models integrated with action chunking. It preserves action performance while eliminating the bottlenecks in the efficiency of autoregressive decoding.
\item We design a decoding-process-only acceleration strategy for VLA inference.
It enables friendly deployment on VLA models and seamlessly synergizes with other acceleration methods. 
\item We conduct comprehensive empirical validation across simulation and real-world platforms, with ablation studies characterizing performance tradeoffs.
\end{itemize}

\section{RELATED WORKS}
\label{sec:rela}
\subsection{Vision-Language-Action Models}
Vision-language-action (VLA) models are designed to process both visual feedback from robotic systems and natural language operation instructions as input, generating executable commands for robots. Several large-scale VLA models~\cite{rt2, kim24openvla, quarvla, li2024roboflamingo} have been developed by fine-tuning pre-trained multimodal large models, which inherently possess strong visual question answering (VQA) capabilities, on extensive robot datasets. 
These methods have shown strong performance in both simulated and real-world tasks.
However, the inference speed of VLA models with a large number of parameters is relatively slow, which prevents them from achieving high control frequencies and further limits the consistency of their actions and their effectiveness when learning flexible tasks from high-frequency demonstrations~\cite{fast}. 
This paper aims to improve inference speed, thereby partially alleviating the aforementioned issues.

\subsection{Action Chunking}
Predicting and executing a sequence of actions without intermediate replanning, which is known as action chunking, is increasingly used in robot learning from human demonstrations.
This approach involves two key strategies. First, it predicts multi-step action sequences and executes them either fully or partially~\cite{act,chi2023diffusion}. Second, it models the distribution of action chunks and performs sampling from the learned model, either independently~\cite{chi2023diffusion,consistencypolicy} or with weak dependencies~\cite{janner2022planning,act}, to facilitate sequential decision-making. While some research highlights the effectiveness of this method in achieving high-performing policies in laboratory settings ~\cite{act,chi2023diffusion}, other studies report contrasting results in real-world applications~\cite{lee2024behavior}. 
Further, \cite{bid}~analyzed the different outcomes under practical conditions and proposed a bidirectional decoding to balance consistency and reactivity.
One of the state-of-the-art VLA model, pi0 \cite{black2024pi_0}, use an action chunking architecture with flow matching to represent complex continuous action distributions.
It validates the effectiveness of action chunking in VLA models.
In this paper, we aim to tackle a significant problem existing in the VLA models with action chunking that the inference speed is severely limited.

\begin{figure*}[t]
    \centering
    \includegraphics[width=0.95\textwidth]{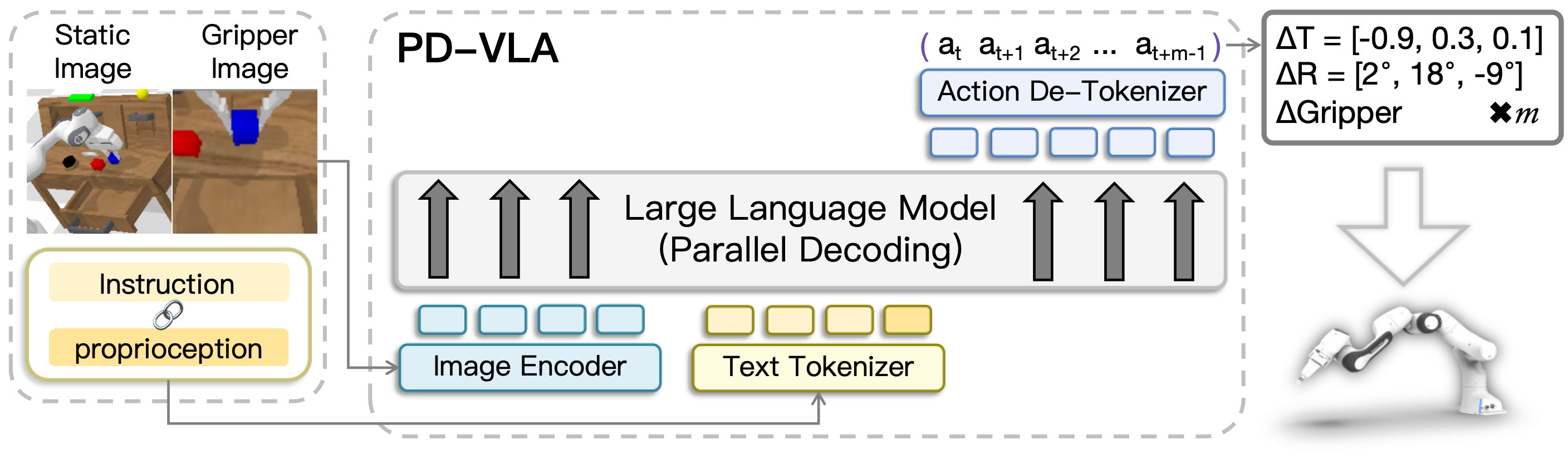}
    \captionsetup{font=footnotesize}
    \caption{The network architecture of our \method~with a chunk size of $m$. Given images, proprioception and language instructions, our method first tokenizes the input and then feeds the results into the LLM in a parallel decoding manner. The LLM outputs action tokens, which are finally detokenized into valid action values and deployed on the mechanical arm.}
    \label{fig:architecture}
\end{figure*}

\subsection{Acceleration for Vision-Language-Action Models}
Various acceleration strategies, including quantization~\cite{lin2024awq} and token pruning~\cite{fastv}, have been effectively applied to LLMs, yet they often fail to meet the stringent real-time requirements of action generation. 
Efforts to enhance efficiency have led to architectural modifications in VLA models, such as DeeR-VLA~\cite{DeeR-VLA}, which dynamically adjusts inference depth, and QAIL~\cite{park2024quantization}, which integrates quantization-aware training. Further innovations, like RoboMamba~\cite{liu2024robomamba} and TinyVLA~\cite{wen2024tinyvla}, replace traditional attention mechanisms or focus on developing lightweight models from the ground up, frequently necessitating model re-training and additional data collection. Meanwhile, VLA-Cache~\cite{vlacache} selectively caches static tokens and recomputes only dynamic or task-relevant ones. 
FAST~\cite{fast} proposes a compression-based tokenization scheme based on the discrete cosine transform.
In contrast, our \method~enhances inference speed by optimizing the decoding mechanism, offering a more practical and deployment-friendly solution compared to the above methods, as shown in Table~\ref{tab:1}.

\section{METHOD}

In this section, we introduce the details of our method \method. We first present the architecture of our VLA model in subsection~\ref{sec:3.1}. Subsequently, we incorporate action chunking with our VLA model in subsection~\ref{sec:3.2}.
Finally, we present parallel decoding to accelerate inference in subsection~\ref{sec:3.3}.

\subsection{Vision-language-action Model}
\label{sec:3.1}
\noindent
\textbf{Model Architecture.}
We build a fundamental VLA model, LLaVA-VLA, on the widely recognized vision-language model, LLaVA~\cite{llava}, ensuring a generalizable and comprehensive exploration.
LLaVA mainly consists of a large language model $\textnormal{LLM}$ and a vision encoder $f_{\textnormal{encoder}}$.
It takes two images as input, a static image $I_{\textnormal{static}}$ and a gripper image $I_{\textnormal{gripper}}$, to get a comprehensive observation. 
Then the images are processed through $f_{\textnormal{encoder}}$ into the visual tokens $h_{I}$.
Along with the input images, the text instructions and proprioceptive input  are first concatenated into a unified instruction $S$, which is then tokenized into tokens $h_S$ via a tokenizer $T$.
Then the $\textnormal{LLM}$ takes in text tokens $h_S$ and image tokens $h_I$ and autoregressively generates action tokens $h_\textnormal{act}$.
Finally, the action tokens are detokenized into 7-dimensionl action $a$.
The whole process can be formulated as:
\begin{equation}
\label{eq:1}
\begin{aligned}
a &= {\rm Detokenize}(h_{\textnormal{act}})={\rm Detokenize}(\textnormal{LLM}(h_I, h_S))\\
  &= {\rm Detokenize}(f_\textnormal{encoder}(I_\textnormal{static}, I_\textnormal{gripper}),T(S)),
  \end{aligned}
\end{equation}

\noindent
\textbf{Action Tokenization.}
Here, we discretize a continuous action $a$ into 256 uniformly spaced bins and represent them as integer indices. Specifically, we utilize the 256 least frequently used tokens in the language model vocabulary to serve as action tokens $h_\textnormal{act}$. Therefore, the robot action tokens across all dimensions can be concatenated with a space character to form a textual string, which serves as the training label. Consequently, a 7-dimensional action $a$ is formatted as:
\begin{equation}
a = [X, \; Y\, \; Z, \; \phi, \; \theta, \; \psi, \; G],
\end{equation}
where $X,Y,Z$ represent the Cartesian coordinates of the end effector's position, $\phi,\theta,\psi$ denote the rotation angles of the end effector along each axis, and $G$ is the gripper state.

\subsection{Action Chunking for VLA Models} 
\label{sec:3.2}
Based on the above fundamental VLA model, we incorporate the action chunking~\cite{act} techniques.
Recent works have pursued a generative approach equipped with action chunking, which predicts a sequence of actions over multiple time steps and executes all or part of the sequence~\cite{act, chi2023diffusion, bid}.
Some studies find this approach improves manipulation performance and execution inference in imitation learning~\cite{act}, diffusion policies~\cite{chi2023diffusion, consistencypolicy}, and VLA models~\cite{vlas}.
Action chunking allows the learner to better capture temporal dependencies in demonstrations and generate more consistent and stable actions~\cite{bid}. 
We integrate action chunking with the VLA model by extending the effective action horizon (chunk size). 
At the current time step $t$, given chunk size $m$, the predicted actions will be extended into an action sequences $\mathcal{A_t}=[{a_t,a_{t+1},a_{t+2},...,a_{t+m-1}}]$, where each element is defined in Equation~(\ref{eq:1}).
Here, following previous work~\cite{vlas}, we set the chunk size to $5$.

However, extended action sequences consume longer single inference time, which impacts the continuity and effectiveness of the actions. 
Therefore, there is an urgent need to propose a more efficient action decoding method.

\subsection{Parallel Decoding for VLA Models}
\label{sec:3.3}
To meet the demands of a more efficient decoding algorithm, we propose parallel decoding for VLA models integrated with action chunking.
In this subsection, we first revisit the theory of AR decoding. 
Then, by leveraging Jacobi decoding, we break the sequential dependency to achieve parallel decoding, and further analyze and refine the approach based on the structural characteristics of VLA. 
Finally, we analyze the acceleration phenomenon demonstrated by parallel decoding.

\noindent
\textbf{Preliminary: Jacobi Decoding.}
Given a prompt $\bm{x}$, comprising both textual and visual components, and a pre-trained LLM model $p(\cdot| \bm{x})$, we typically predicte tokens using the standard AR decoding method under a greedy strategy, \textit{i.e.},
\begin{equation}
\label{eq:ar_decoding}
\begin{aligned}
y_i = \underset{y}{\mathrm{arg\,max}}\ p(y | \mathcal{Y}_i, \bm{x}) \;\, \text{for}\,\, i = 1,\dots,n
\end{aligned}
\end{equation}
where $\mathcal{Y}_i$ denotes $\{y_{1},  \ldots, y_{i-1} \}$, $n$ denotes the decoding horizon, representing the number of tokens to predict.
As shown in Fig.~\ref{fig:teaser}, $n$ forward passes of the LLM are required to obtain $n$ tokens $\mathcal{Y}_n$. 
The sequential characteristic in AR decoding restricts the efficient generation of a lengthy token sequence. 

Compared with the aforementioned AR decoding, Jacobi decoding \cite{jacobidecoding, cllm} has shown the capacity to tackle lengthy token sequences.
Concretely, supposing $f(y_i, \mathcal{Y}_i, \bm{x}):= y_i-{\mathrm{arg\,max}_y}\ p(y | \mathcal{Y}_i, \bm{x})$, Jacobi decoding re-frames the inference process of LLM in \cref{eq:ar_decoding} as solving a system of nonlinear equations with respect to $y_i$:
\begin{equation}
f(y_i, \mathcal{Y}_i, \bm{x}) = 0 \;\, \text{for}\,\, i = 1,\dots,n. 
\label{eq:y_i}
\end{equation}
There are $n$ unknown parameters~$y_i$ in the nonlinear equation system including $n$ Equation~(\ref{eq:y_i}).
Considering Equation~\ref{eq:ar_decoding}, the system of nonlinear equation system can be formulated as:
\begin{equation}
\label{eq:jacobi_decoding}
\begin{aligned}
\begin{cases}
y_{1}^{(j+1)} &= \underset{y}{\mathrm{arg\,max}}\ p(y | \bm{x}) \\
y_{2}^{(j+1)} &= \underset{y}{\mathrm{arg\,max}}\ p(y | \mathcal{Y}_{1}^{(j)}, \bm{x}) \\
& \vdots \\
y_{n}^{(j+1)} &= \underset{y}{\mathrm{arg\,max}}\ p(y | \mathcal{Y}_{n}^{(j)}, \bm{x}),
\end{cases}
\end{aligned}
\end{equation}
which can be solved in the Jacobi fix-point iteration method~\cite{ortega2000iterative} by using a causal attention mask.

\noindent
\textbf{Our Jacobi Decoding-based Acceleration.} 
In this part, we will introduce how we apply the above Jacobi Decoding to the VLA model.
We first randomly initialize an action token sequence of equal length to the decoding horizon $n$. Both the prompt $x$ and the initialized action sequence $\mathcal{Y}^{(0)}=\{y_{1}^{(0)},  \ldots, y_{n}^{(0)} \}$ are fed into the VLA model simultaneously.
To break the sequential dependencies in the conventional VLA model, we replace the above causal attention mechanism with a bidirectional attention mechanism, which re-formulate the system of nonlinear equations Equation~\ref{eq:jacobi_decoding} as:
\begin{equation}
\label{eq:pdvla}
\begin{aligned}
\begin{cases}
y_{1}^{(j+1)} &= \underset{y}{\mathrm{arg\,max}}\ p(y | \mathcal{Y}^{(j)}, \bm{x}) \\
y_{2}^{(j+1)} &= \underset{y}{\mathrm{arg\,max}}\ p(y | \mathcal{Y}^{(j)}, \bm{x}) \\
& \vdots \\
y_{n}^{(j+1)} &= \underset{y}{\mathrm{arg\,max}}\ p(y | \mathcal{Y}^{(j)}, \bm{x}).
\end{cases}
\end{aligned}
\end{equation}
This enables updates of all action tokens in every single iteration.
The iterations terminate at the step $k$ where $\mathcal{Y}^{(k)}=\mathcal{Y}^{(k-1)}$, and the $\mathcal{Y}^*:=\mathcal{Y}^{(k)}$ is defined as the fixed point. 
The acceleration achieved by Jacobi decoding originates from its ability to predict multiple tokens in the $n$-token sequence in parallel during each forward pass. 
Therefore, the total number of updating iterations $k$ can be smaller than AR decoding, i.e., $k \leq n$.

\begin{table*}[t]
\centering
\small
\captionsetup{font=footnotesize}
\caption{Comparison with various manipulation baselines in terms of success rate and average length.
} 
\label{tab:baselines}
\scalebox{0.98}{
\begin{tabular}{ccccccccc}
\toprule
\multirow{2}{*}{Method} & \multirow{2}{*}{Input} & \multirow{2}{*}{Data}  &
\multicolumn{5}{c}{Success Rate (\%)} & Avg. len. \\
& & &  1/5 & 2/5 & 3/5& 4/5& 5/5 & ABCD$\rightarrow$D \\
\midrule
MCIL~\cite{calvin}~\textcolor{gray} & RGB & ALL & 37.3& 2.7& 0.2& 0.0& 0.0 & 0.40 \\
HULC~\cite{HULC}~\textcolor{gray} & RGB & ALL  & 89.2& 70.1& 54.8& 42.0& 33.5& 2.90 \\
RT-1~\cite{rt1}~\textcolor{gray} & RGB & LANG  & 84.4& 61.7& 43.8& 32.3 & 22.7& 2.45 \\
LLaVA-VLA & RGB & LANG & 72.0 & 29.0 & 12.0 & 6.0 & 1.9& 1.20\\
\midrule
\method~(ours) & RGB & LANG & \textbf{94.1} & \textbf{80.0} & \textbf{68.3} & \textbf{61.4} & \textbf{50.5}  & \textbf{3.54} \\
\bottomrule
\end{tabular}
}
\end{table*}
\begin{table*}[t]
\centering
\small
\captionsetup{font=footnotesize}
\caption{
\textbf{Ablation study}. We ablate 2 core components of our methods, action chunking (AC) and parallel decoding (PD), to emphasize their significance.
In addition, we replace PD with other acceleration methods. 
Here, we select 2 state-of-the-art training-free methods for VLM.
} 
\label{tab:ablation}
\begin{tabular}{ccccccccc}
\toprule
\multirow{2}{*}{Method} & \multicolumn{5}{c}{Success Rate (\%)} & Avg. len. &Avg. Speed & Frequency\\
& 1/5 & 2/5 & 3/5& 4/5& 5/5 & ABCD$\rightarrow$D &(Token/s) &(Hz) \\
\midrule

LLaVA-VLA & 72.0 & 29.0 & 12.0 & 6.0 & 1.9& 1.20 & 39.56 & 1.81\\
\midrule
w/o AC  & 71.0 & 25.0 & 8.0 & 6.0 & 2.0& 1.12 & 39.86 & 1.82\\
w/o PD  & 91.8 & 82.4 & 71.0 & 62.8 & 52.6& 3.61 & 41.44 & 3.60 \\
w/o PD, w/ FastV  & 90.1 & 77.2 & 62.4 & 55.4 & 46.5 & 3.31 & 28.69 & 2.54\\
w/o PD, w/ SparseVLM  & 83.2 & 63.2 & 46.0 & 36.0 & 26.4 & 2.55 & 32.43 & 2.83\\
\midrule
\method~(ours)  & 94.1 & 80.0 & 68.3 & 61.4 & 50.5  & 3.54 & 52.84 & 4.56\\
\bottomrule
\end{tabular}
\end{table*}

\begin{table*}[t]
\small
\centering
\caption{\textbf{Evaluation and comparison on the LIBERO benchmark.}}
\begin{tabular}{lccccc}
\toprule
\textbf{Method} & \textbf{Spatial} & \textbf{Object} & \textbf{Goal} & \textbf{Long} & \textbf{Average} \\
\midrule
Diffusion Policy~\cite{chi2023diffusion}         & 78.3\% & 92.5\% & 68.3\% & 50.5\% & 72.4\% \\
Octo~\cite{octo_2023}               & 78.9\% & 85.7\% & 84.6\% & 51.1\% & 75.1\% \\
OpenVLA~\cite{kim2024openvla}       & 84.9\% & 88.4\% & 79.2\% & 53.7\% & 76.5\% \\
SpatialVLA~\cite{qu2025spatialvla}  & 88.2\% & 89.9\% & 78.6\% & 55.5\% & 78.1\% \\
CoT-VLA~\cite{zhao2025cot}          & 87.5\% & 91.6\% & 87.6\% & 69.0\% & 81.1\%\\
WorldVLA~\cite{worldvla}          & 87.6\% &  96.2\% & 83.4\% & 60.0\% & 81.8\%\\
ThinkAct~\cite{huang2025thinkact}          & 88.3\% &  91.4\% & 87.1\% & 70.9\% & 84.4\%\\
$\pi_0$-FAST~\cite{pertsch2025fast} & 96.4\% &96.8\% & 88.6\% & 60.2\% & 85.5\% \\
MolmoAct~\cite{lee2025molmoact}          & 87.0\% &  95.4\% & 87.6\% & 77.2\% & 86.6\%\\
FlowVLA~\cite{zhong2025flowvla}          & 93.2\% &  95.0\% & 91.6\% & 72.6\% & 88.1\%\\
DreamVLA~\cite{dreamvla25}          & \textbf{97.5\%} &  94.0\% & 89.5\% & 89.5\% & 92.6\%\\
$\pi_0$~\cite{black2024pi_0} & 96.8\% & \textbf{98.8\%} & \textbf{95.8\%} & 85.2\% & 94.2\% \\
\midrule
\method~(ours) & 95.5\% & 96.7\% & 94.9\% & \textbf{91.7\%} & \textbf{94.7\%}\\
\bottomrule
\end{tabular}
\label{tab:libero}
\end{table*}

Different decoding horizons $n$ result in different parallel decoding patterns, which further influence the effectiveness.
When $n$ is less than the total action dimensions $l$, it decodes $n$ action token in one iteration and then proceeds to the next $n$ token, until covering the total $l$ token.
In mathematics, it equals several Jacobi decodings with several Gauss-Seidel steps.
The decoding horizon $n$ is often selected as powers of $2$.
For a VLA model with a chunk size of $m$, the response length is $l=7m+2$, with the addition of a blank beginning token and an ending token. 
However, when the value of $l$ is not powers of $2$, some redundant tokens may be predicted.
Considering the structural properties of the VLA task, setting the per-action dimension as the value of $n$ is a feasible approach to enable the model to better learn the physical meaning of action.
A simple but effective idea is to set $n=l$, which enables inference finished in a single Jacobi decoding.
This configuration is more conducive to enabling the model to inherit the action modeling capabilities of the original distribution.
Following the analysis above, we select 7, 16, and 37 as the value of $n$, to better analyze the effectiveness of different decoding horizons.
In the following, we will analyze an acceleration phenomenon caused by parallel decoding.
During decoding, our \method~exhibits the capability of predicting correct action tokens preemptively, even with preceding incorrect tokens, while ensuring the tokens remain unchanged. 
We term such tokens as \textit{fixed tokens}, whose existence allows simultaneous extension of discontinuous correct tokens within the $n$-token sequence. 
In particular, for VLA models, the token denoting gripper opening only has two values, $0$ for close and $1$ for open, which tokens are easier to predict as \textit{fixed tokens}.
This phenomenon contributes to the fast convergence in parallel decoding, thereby leading to a considerable generation speedup.

\section{EXPERIMENTS}
We concentrate on several experiments to answer the following questions: 
\textbf{Q1.} How does the effectiveness of \method~compare with baselines and other acceleration methods?
\textbf{Q2.} Is the coordination among different components effective?
\textbf{Q3.} How does the acceleration phenomenon vary across different decoding horizons? 
\textbf{Q4.} Can \method~be effectively deployed in real-world robotic systems?

\subsection{Experiments Setup}
\noindent
\textbf{Simulation environment.}
The CALVIN benchmark~\cite{calvin} is built on top of the PyBullet~\cite{pybullet} simulator and involves a Franka Panda Robot arm that manipulates the scene. 
CALVIN consists of 34 tasks and 4 different environments (A, B, C and D). 
We evaluate all methods on the classic CALVIN ABCD$\rightarrow$D setup~\cite{calvin}. 
We report the success rate and the average number of completed sequential tasks.
LIBERO~\cite{liu2023libero} is another widely-used simulated manipulation benchmark with 4 suites (Spatial, Object, Goal, Long). 
We report the success rate for each evaluation suite as well as the overall average, where each suite comprises 10 tasks, and each task is evaluated over 50 rollouts.

\noindent
\textbf{Evaluate metrics.}
The CALVIN long-horizon challenge is a sequential task comprising five subtasks. 
We report the success rates for each subtask and the average completed length across all five tasks. 
To quantitatively evaluate inference speed, we introduce inference speed (measured in tokens per second). 
Additionally, considering the requirements of robotic tasks, we also report execution frequency (in Hertz, Hz).

\noindent
\textbf{Implementation details.}
In this paper, we use vicuna-7b-v1.5~\cite{vicuna} as the LLM backbone and clip-vit-large-patch14-336~\cite{clip} as the vision encoder to build LLaVA-7b-v1.5~\cite{llava}.

\noindent
\textbf{Training details.}
Our fundamental model LLaVA-VLA is trained using 8 NVIDIA H100 GPUs over 1 epoch, which requires approximately 10 hours. Notably, our \method~does not incur extra training costs.

\subsection{Results on CALVIN Benchmark}
\noindent
\textbf{Comparison with other models.}
In Table~\ref{tab:baselines}, we benchmark our method against several representative models. For a comprehensive comparison, we include various baselines, such as the official MCIL~\cite{calvin} model and other prevalent models like HULC~\cite{HULC} and RT-1~\cite{rt1}. Our method demonstrates competitive performance, with \method~achieving significant improvements over the fundamental LLaVA-VLA model, further validating its effectiveness.

\noindent
\textbf{Comparison with other acceleration methods for VLA models.}
For fair comparisons, we deployed 2 state-of-the-art VLM acceleration methods, FastV~\cite{fastv} and SparseVLM~\cite{zhang2024sparsevlm}, on the traditional VLA model with action chunking.
However, neither FastV nor SparseVLM really improves the inference speed. 
While FastV largely preserves manipulation performance, masking tokens in attention computation incurs additional overhead, leading to slower inference speeds.
SparseVLM witnesses decrease both in success rates and inference speed because it incurs extra costs from token pruning, merging, and recycling.

\subsection{Results on LIBERO Benchmark}
We further validate the superior performance of PD-VLA on the LIBERO benchmark. Notably, this variant corresponds to our method integrated with an action head, following the OpenVLA-OFT~\cite{kim2025fine} setting. Compared with prior state-of-the-art approaches, PD-VLA achieves the best average performance, attaining a 91.7\% success rate on the most challenging LIBERO-Long benchmark.

\subsection{Ablation Study}
Table~\ref{tab:ablation} presents a detailed summary of the ablation studies performed on two key components of our \method. 
These components enable \method~to improve 2.34 in success rates and realize 2.52$\times$ execution frequency compared to the fundamental model LLaVA-VLA.
The ablation findings are as follows:

First, the ablation study of the action chunking demonstrates a significant performance boost with its inclusion. 
By extending chunk sizes, the consistency and stability is improved, showing 2.42 improvements in average length. 
However, the decoding speed is still limited, resulting in a longer single inference time.
By reducing inference counts, the execution frequency is improved 2.51$\times$.

Second, the ablation study of parallel decoding reveals the inefficiency in the inference process.
Parallel decoding substantially increases the average decoding speed by 1.28$\times$, thus the single inference time is reduced and satisfies the demand of high-frequency inference.

The above two key components complement each other: action chunking enhances action consistency while improving execution frequency, whereas parallel decoding mitigates inference inefficiency by accelerating the decoding process.
Together, they strike a balance between performance and high-frequency inference.

\begin{figure*}[t!]
    \centering
    \includegraphics[width=0.995\textwidth]{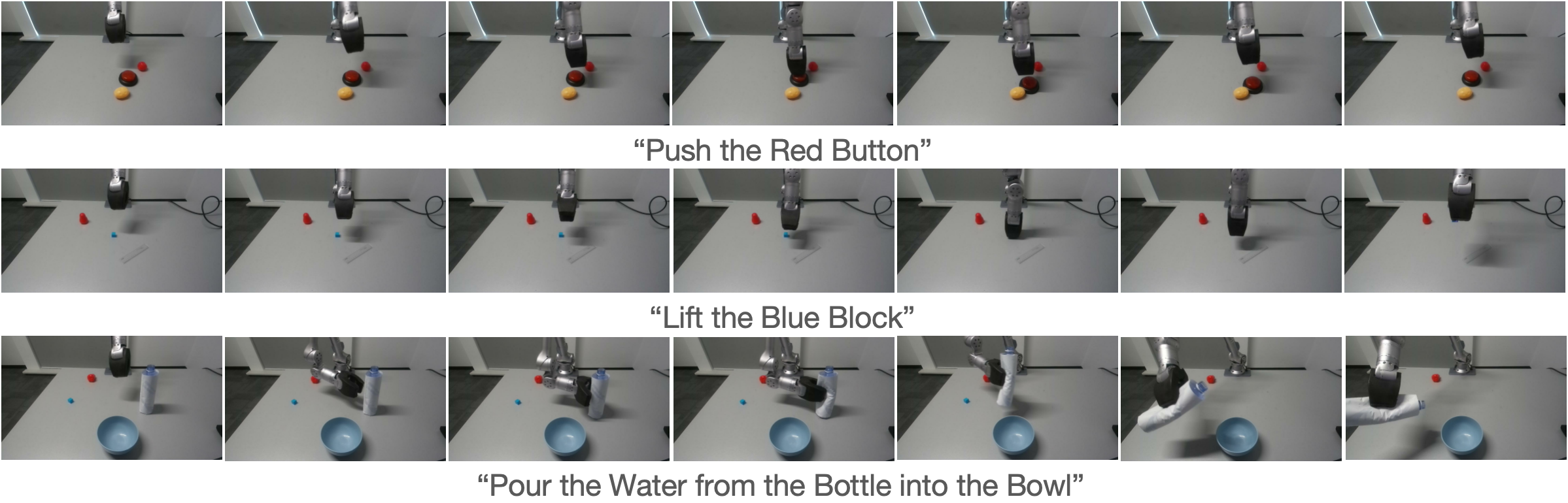}
    \captionsetup{font=footnotesize}
    \caption{\textbf{Representative results of real-world experiments}. The sequential images showcase the trajectories of a robotic arm successfully executing three tasks.}
    \label{fig:real}
\end{figure*}

\subsection{Decoding Horizon and Acceleration Phenomenon}
We further conduct an in-depth investigation into the differences in acceleration phenomenon across various decoding horizons and their impact on performance.
We compare methods with different decoding horizons in Table~\ref{tab:mechanism} on the numbers of $fixed$ tokens, average length, decoding speed, and execution frequency.
The method with a decoding horizon of $37$ shows the strongest manipulation abilities with the highest decoding speed.
This setting ensures the inheritance of the modeling of the original action distribution by predicting the whole action sequences together.
The $7$-token method performs better than the $16$-token one because it aligns with the distribution of the single action, facilitating more efficient decoding in accordance with the action structure.
With the increasing decoding horizon, the number of \textit{fixed} tokens increases accordingly, which contributes to the decoding speed improved from 41.48 to 52.84 tokens/second. 
Notably, the redundant tokens when $n=16$ make execution frequency even lower. 
\begin{table}[t]
\centering
\small
\captionsetup{font=footnotesize}
\caption{Analysis of the different decoding horizons and acceleration phenomenon between them. } 
\label{tab:mechanism}
\begin{tabular}{cccccc}
\toprule
decoding  &fixed & Average &Avg. Speed & Frequency\\
horizon & token count & length &(Token/s) &(Hz) \\
\midrule
7  &5.17 & 3.24 & 41.48 & 3.60   \\
16 &6.75 & 3.19 & 48.74 & 3.25 \\
37 &\textbf{8.75} & \textbf{3.64} & \textbf{52.84} & \textbf{4.56} \\
\bottomrule
\end{tabular}
\end{table}

\begin{figure}[t]
    \centering    \includegraphics[width=0.49\textwidth]{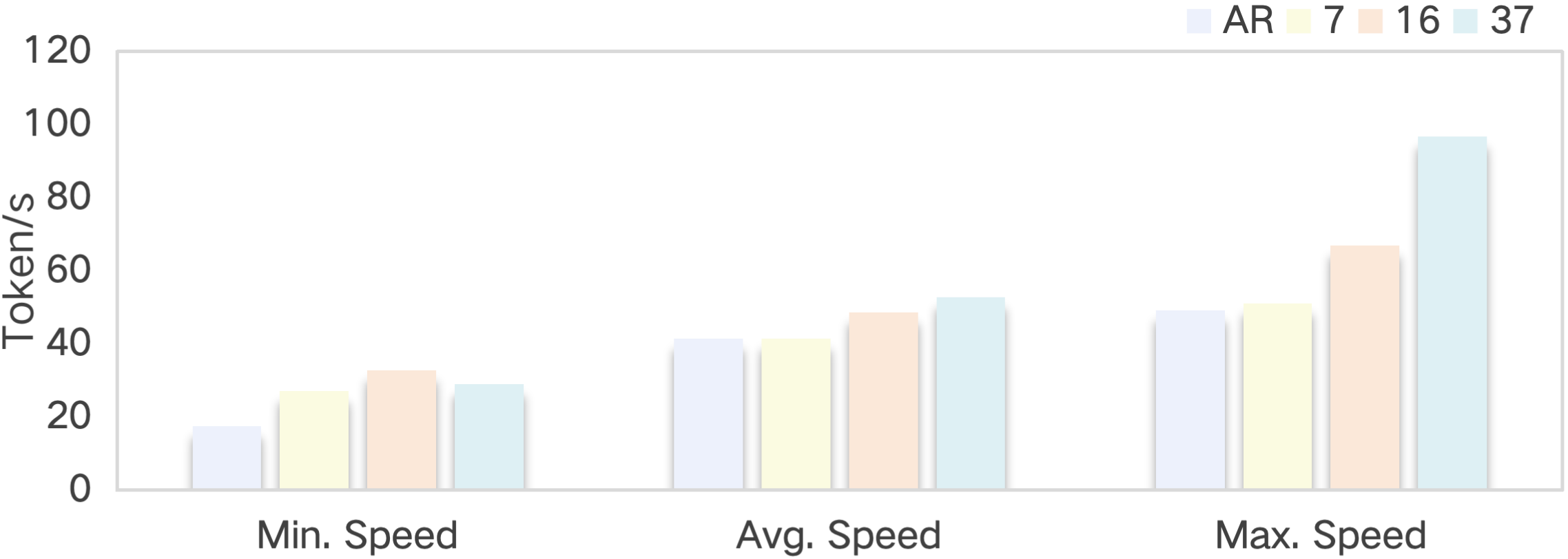}
    \captionsetup{font=footnotesize}
    \caption{Comparison of minimum, average, and maximum inference speed (tokens per second) between AR decoding and parallel decoding with different decoding horizons $n$.}
    \label{fig:speed}
\end{figure}
Fig.~\ref{fig:speed} illustrates the speed distribution of different decoding horizons.
We observe a remarkable increase in maximum speed as $n$ grows. At $n=37$, the maximum speed reaches approximately twice that of $n=7$ and AR, thanks to the reduction of the number of iterations. 
This finding highlights the potential of parallel decoding to achieve greater acceleration.

\begin{figure}[t]
    \centering    \includegraphics[width=0.49\textwidth]{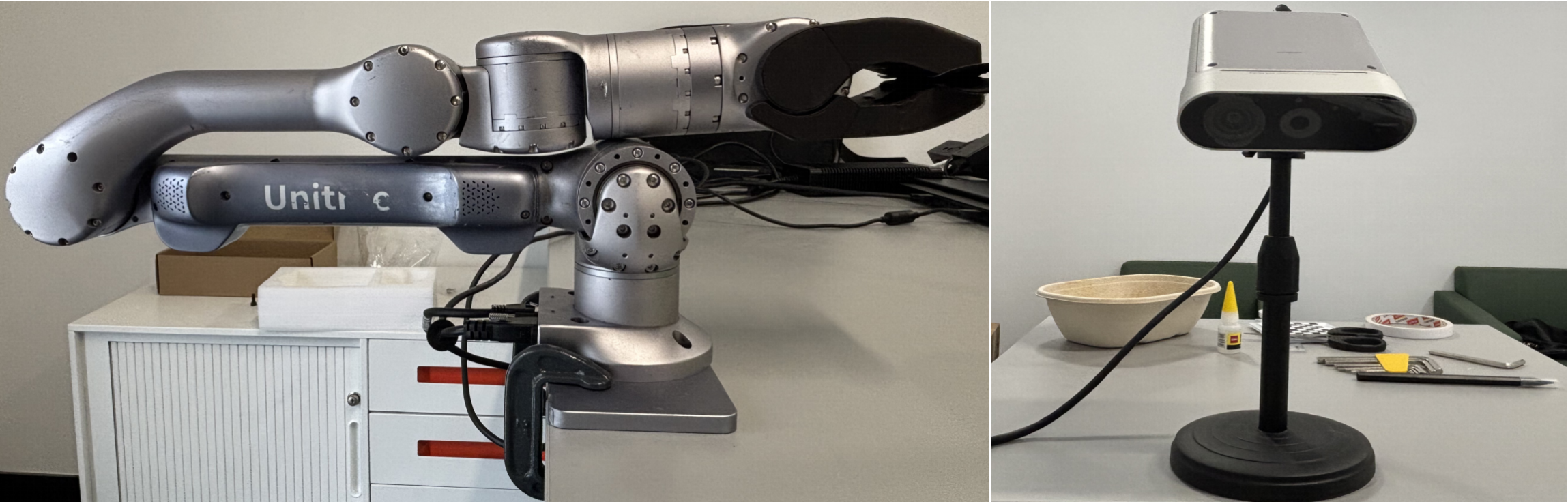}
    \captionsetup{font=footnotesize}
    \caption{Real-World Setup. The left panel shows the mechanical arm and the right panel shows the camera used.}
    \label{fig:setup}
\end{figure}

\subsection{Real-world Experiments}
\noindent
\textbf{System set-up.} 
The real-world settings are shown in Fig.~\ref{fig:setup} We set up real-world experiments based on a 6-DOF Unitree Z1-Pro mechanical arm with a 1-DOF gripper. We provide images using an ORBBEC Femto Mega camera at a front view.
We collect a small robotic dataset including 3 tasks: push the button, lift the block, and pour the water into the bowl.
Each task contains 50 demonstrations and evaluates 10 episodes for success rates.

\begin{table}[t]
\centering
\small
\captionsetup{font=footnotesize}
\caption{Comparison with LLaVA-VLA in the real world. We report success rates as metrics.} 
\label{tab:real}
\scalebox{0.98}{
\begin{tabular}{cccc}
\toprule
Method & ``push button''  & ``lift block'' & ``pour water''\\
\midrule
LLaVA-VLA& 60\% &40\% & 10\%    \\
\method & 80\% & 70\% & 60\%  \\
\bottomrule
\end{tabular}
}
\end{table}

\noindent
\textbf{Quantative Results.} Table~\ref{tab:real} shows that our \method~got higher success rates compared with LLaVA-VLA.
Benefiting from the strong visual generalization capabilities of the pre-trained VLM, both LLaVA-VLA and \method~ successfully accomplish ``push button" and ``lift block". However, thanks to the combination of action chunking and parallel decoding, \method~can produce more consistent actions, resulting in 20\% and 30\% improvements in success rates, respectively.
For the task ``pour water", LLaVA-VLA failed to complete this task, while \method~has a 50\% higher success rate.
This task challenges the flexibility and manipulation abilities of models, \method~has a higher execution frequency and adjusts the action according to the real-time image.

\noindent
\textbf{Visualization.}
Fig.~\ref{fig:real} presents visualizations of real-world experiments on three tasks. 
All tasks include distractors to validate the robustness of the model. 
In the ``push button" task, the model successfully identifies the red button and moves the end-effector to press it. 
In the ``lift block" task, the model accurately recognizes the small blue cube, precisely positions the end-effector, opens it for a firm grasp, and then lifts the robotic arm. 
The ``pour water" task requires more dexterous manipulation, as it involves a non-flexible end-effector grasping a non-rigid plastic bottle and tilting it to pour water into a bowl. Any inconsistency during the grasping process could easily lead to the bottle being dropped. However, \method~demonstrates smooth and consistent actions throughout the process, successfully completing the task. This demonstrates its suitability for real-time robotic applications.

\section{CONCLUSION}
This paper analyzes the inefficiency of autoregressive VLA  models integrated with action chunking.
Therefore, we propose \method, which is a novel parallel decoding method designed for the VLA model integrated with action chunking.
Instead of predicting each action token sequentially, our \method~tries to predict every token simultaneously in several iterations, thus hugely improving the decoding efficiency.
Benefiting from parallel decoding and action chunking, the model strikes a balance between performance and high-frequency inference.
Extensive experiments demonstrate that our \method~significantly improves inference speeds and execution frequency while maintaining competitive success rates.
Real-world experiments validate the effectiveness of \method~in the real world.

In the future, we will focus on optimizing the decoding algorithm and model to prevent redundant iteration processes during parallel decoding, thereby enabling faster convergence to a fixed point.

\section{Acknowledgement}
We sincerely thank Wei Li and Jie He for their valuable assistance in implementing the variant equipped with the action head.

\bibliographystyle{IEEEtran}
\bibliography{root}



\end{document}